\documentclass[10pt,twocolumn,letterpaper]{article}

\usepackage{cvpr}
\usepackage{times}
\usepackage{epsfig}
\usepackage{graphicx}
\usepackage{amsmath}
\usepackage{amssymb}
\usepackage{booktabs}
\usepackage{multirow}
\usepackage{paralist}

\usepackage[pagebackref=true,breaklinks=true,letterpaper=true,colorlinks,bookmarks=false]{hyperref}

\cvprfinalcopy 

\def\cvprPaperID{****}

\DeclareMathOperator*{\argmax}{\arg\!\max}
\DeclareMathOperator*{\argmin}{\arg\!\min}

\newcommand{\opt}[1]{\tilde{#1}}
\newcommand{\gt}[1]{{#1}^*}

\newcommand{\crd}{\mathbf{y}}
\newcommand{\crds}{Y}

\newcommand{\pnp}{\mathbf{H}}
\newcommand{\mdl}{\mathbf{h}}
\newcommand{\sam}[1]{{#1}_{\mathrm{SoftAM}}}
\newcommand{\am}[1]{{#1}_{\mathrm{AM}}}
\newcommand{\smpl}[1]{{#1}_{\mathrm{DSAC}}}

\newcommand{\refine}{\mathbf{R}}
\newcommand{\loss}{\ell}

\newcommand{\paramv}{\mathbf{v}}
\newcommand{\paramw}{\mathbf{w}}

\newcommand{\expectation}[2]{\mathbb{E}_{#1}\left[ #2 \right]}
\newcommand{\derv}[1]{\frac{\partial}{\partial #1}}

\begin{document}

\title{DSAC - Differentiable RANSAC for Camera Localization}

\author{Eric Brachmann$^1$, Alexander Krull$^1$, Sebastian Nowozin$^2$\\
Jamie Shotton$^2$, Frank Michel$^1$, Stefan Gumhold$^1$, Carsten Rother$^1$\\
$^1$ TU Dresden, $^2$ Microsoft
}

\maketitle

\begin{abstract}

RANSAC is an important algorithm in robust optimization and a central building block for many computer vision applications. In recent years, traditionally hand-crafted pipelines have been replaced by deep learning pipelines, which can be trained in an end-to-end fashion. However, RANSAC has so far not been used as part of such deep learning pipelines, because its hypothesis selection procedure is non-differentiable. In this work, we present two different ways to overcome this limitation. The most promising approach is inspired by reinforcement learning, namely to replace the deterministic hypothesis selection by a probabilistic selection for which we can derive the expected loss \wrt to all learnable parameters. We call this approach DSAC, the differentiable counterpart of RANSAC. We apply DSAC to the problem of camera localization, where deep learning has so far failed to improve on traditional approaches. We demonstrate that by directly minimizing the expected loss of the output camera poses, robustly estimated by RANSAC, we achieve an increase in accuracy. In the future, any deep learning pipeline can use DSAC as a robust optimization component\footnote{Source code and trained models are publicly available:\\ \url{https://hci.iwr.uni-heidelberg.de/vislearn/research/scene-understanding/pose-estimation/\#DSAC}}.

\end{abstract}

\section{Introduction}

Introduced in 1981, the random sample consensus (RANSAC) algorithm \cite{ransac1981} remains the most important algorithm for robust estimation. It is easy to implement, it can be applied to a wide range of problems and it is able to handle data with a substantial percentage of outliers, \ie data points that are not explained by the data model. RANSAC and variants thereof \cite{mlesac2000, preemptive2003, prosac2005} have, for many years, been important tools in computer vision, including multi-view geometry \cite{mutliview2004}, object retrieval \cite{Philbin07}, pose estimation \cite{shotton13scorf, brachmann2014pose6d} and simultaneous localization and mapping (SLAM) \cite{orbslam2015}. Solutions to these diverse tasks often involve a common strategy: Local predictions (\eg feature matches) induce a global model (\eg a homography). In this schema, RANSAC provides robustness to erroneous local predictions. 

Recently, deep learning has been shown to be highly successful at image recognition tasks \cite{Simonyan2014, resnet2015, fastrcnn2015, yolo2015}, and, increasingly, in other domains including geometry \cite{eigen2015, WarpNet2016, kendall2015convolutional, HomographyNet2016}. Part of this recent success is the ability to perform end-to-end training, \ie propagating gradients back through an entire pipeline to allow the direct optimization of a task-specific loss function, examples include \cite{lift2016,netvlad2016,trainDM2016}. 

In this work, we are interested in learning components of a computer vision pipeline that follows the principle: predict locally, fit globally. As explained earlier, RANSAC is an integral component of this wide-spread strategy. We ask the question, whether we can train such a pipeline end-to-end. More specifically, we want to learn parameters of a convolutional neural network (CNN) such that models, fit robustly to its predictions via RANSAC, minimize a task specific loss function.

RANSAC works by first creating multiple model hypotheses from small, random subsets of data points. Then it scores each hypothesis by determining its consensus with all data points. Finally, RANSAC selects the hypothesis with the highest consensus as the final output. Unfortunately, this hypothesis selection is non-differentiable, meaning that it cannot directly be used in an end-to-end-trained deep learning pipeline. 

A common approach within the deep learning community is to soften non-differentiable operators, \eg $\argmax$ in LIFT \cite{lift2016} or visual word assignment in NetVLAD \cite{netvlad2016}. In the case of RANSAC, the non-differentiable operator is the $\argmax$ operator which selects the highest scoring hypothesis. Similar to \cite{lift2016}, we might substitute the $\argmax$ for a soft $\argmax$, which is a weighted average of arguments \cite{Chapelle2010}. We indeed explore this direction but argue that this substitution changes the underlying principle of RANSAC. Instead of learning how to select a good hypothesis, the pipeline learns a (robust) average of hypotheses. We show experimentally that this approach learns to focus on a narrow selection of hypotheses and is prone to overfitting.

Alternatively, we aim to preserve the hard hypothesis selection but treat it as a probabilistic process. We call this approach DSAC -- Differentiable SAmple Consensus -- our new, differentiable counterpart to RANSAC. DSAC allows us to differentiate the \emph{expected} loss of the pipeline \wrt to all learnable parameters. This technique is well known in reinforcement learning, for stochastic computation problems like policy gradient approaches \cite{compgraph2015}.

To demonstrate the principle, we choose the problem of camera localization: From a single RGB image in a known static scene, we estimate the 6D camera pose (3D translation and 3D rotation) relative to the scene.  We demonstrate an end-to-end trainable solution for this problem, building on the scene coordinate regression forest (SCoRF) approach \cite{shotton13scorf,valentin2015cvpr,brachmann2016}.  The original SCoRF approach uses a regression forest to predict the 3D location of each pixel in an observed image in terms of `scene coordinates'.  A hypothesize-verify-refine RANSAC loop then randomly select scene coordinates of four pixel locations to generate an initial set of camera pose hypotheses, which is then iteratively pruned and refined until a single high-quality pose estimate remains. In contrast to previous SCoRF approaches, we adopt two CNNs for predicting scene coordinates and for scoring hypotheses. More importantly, the key novelty of this work is to replace RANSAC by our new, differentiable DSAC. 

\noindent{\bf Our contributions are in short:}
\begin{compactitem}

\item We present and discuss two alternative ways of making RANSAC differentiable, by soft $\argmax$ and probabilistic selection. We call our new RANSAC version, with the latter option, DSAC (Differentiable SAmple Consensus).

\item  We put both options into a new end-to-end trainable camera localization pipeline. It contains two separate CNNs, linked by our new RANSAC, motivated by previous work \cite{shotton13scorf, krull2015analysis}.

\item We validate experimentally that the option of probabilistic selection is superior, \ie less sensitive to overfitting, for our application. We conjecture that the advantage of probabilistic selection is allowing hard decisions and, at the same time, keeping broad distributions over possible decisions.

\item We exceed the state-of-the-art results on camera localization by 7.3\%.
\end{compactitem}

\subsection{Related Work}
\label{sec:related}

Over the last decades, researchers have proposed many variants of the original RANSAC algorithm \cite{ransac1981}. Most works  focus on either or both of two aspects: speed \cite{Chum2003,preemptive2003, prosac2005}, or quality of the final estimate \cite{mlesac2000, Chum2003}. For detailed information about RANSAC variants we refer the reader to \cite{ransacsurvey2008}. To the best of our knowledge, this work is the first to introduce a differentiable variant of RANSAC for the purpose of end-to-end learning.
In the following, we review previous work on differentiable algorithms and solutions for the problem of camera localization.

\noindent{\bf Differentiable Algorithms.} The success of deep learning began with systems in which a CNN processes an image in one forward pass to directly predict the desired output, \eg class probabilities \cite{alexnet2012}, a semantic segmentation \cite{fcn2015} or depth values and normals \cite{eigen2015}. Given a sufficient amount of training data, CNNs can autonomously discover useful strategies for solving a task at hand, \eg hierarchical part-structures for object recognition \cite{zeiler2014visualizing}. 

However, for many computer vision tasks, useful strategies have been known for a long time. Recently, researchers started to revisit and encode such strategies explicitly in deep learning pipelines. This can reduce the necessary amount of training data compared to CNNs with an unconstrained architecture \cite{semantic2016}. Yi \etal \cite{lift2016} introduced a stack of CNNs that remodels the established sparse feature pipeline of detection, orientation estimation and description, originally proposed in \cite{Lowesift}. Arandjelovic~\etal~ \cite{netvlad2016} mapped the Vector of Locally Aggregated Descriptors (VLAD) \cite{vlad2013} to a CNN architecture for place recognition. Thewlis \etal \cite{trainDM2016} substituted the recursive decoding of Deep Matching \cite{Revaud2016} with reverse convolutions for end-to-end trainable dense image matching.

Similar in spirit to these works, we show how to train an established, RANSAC-based computer vision pipeline in an end-to-end fashion. Instead of substituting hard assignments by soft counterparts as in \cite{lift2016, netvlad2016}, we enable end-to-end learning by turning the hard selection into a probabilistic process. Thus, we are able to calculate gradients to minimize the expectation of the task loss function \cite{compgraph2015}.

\noindent{\bf Camera Localization.} The SCoRF camera localization pipeline \cite{shotton13scorf}, already discussed in the introduction, has been extended in several works. Guzman-Rivera \etal \cite{guzman2014multi} trained a random forest to predict diverse scene coordinates to resolve scene ambiguities. Valentin \etal \cite{valentin2015cvpr} trained the random forest to predict multi-model distributions of scene coordinates for increased pose accuracy. Brachmann~\etal~\cite{brachmann2016} addressed camera localization from an RGB image instead of RGB-D, utilizing the increased predictive power of an auto-context random forest. None of these works support end-to-end learning.

In a system similar to SCoRF but for the task of object pose estimation, Krull \etal \cite{krull2015analysis} trained a CNN to measure hypothesis consensus by comparing rendered and observed images. In this work, we adopt the idea of a CNN measuring hypothesis consensus, but learn it jointly with the scene coordinate regressor and in an end-to-end fashion. 
 
Kendall \etal \cite{kendall2015convolutional} demonstrated that a single CNN is able to directly regress the 6D camera pose given an RGB image, but its accuracy on indoor scenes is inferior to a RGB-based SCoRF pipeline \cite{brachmann2016}.


\section{Method}
\label{sec:method}

\begin{figure*}[h!t]
\includegraphics[scale=0.6]{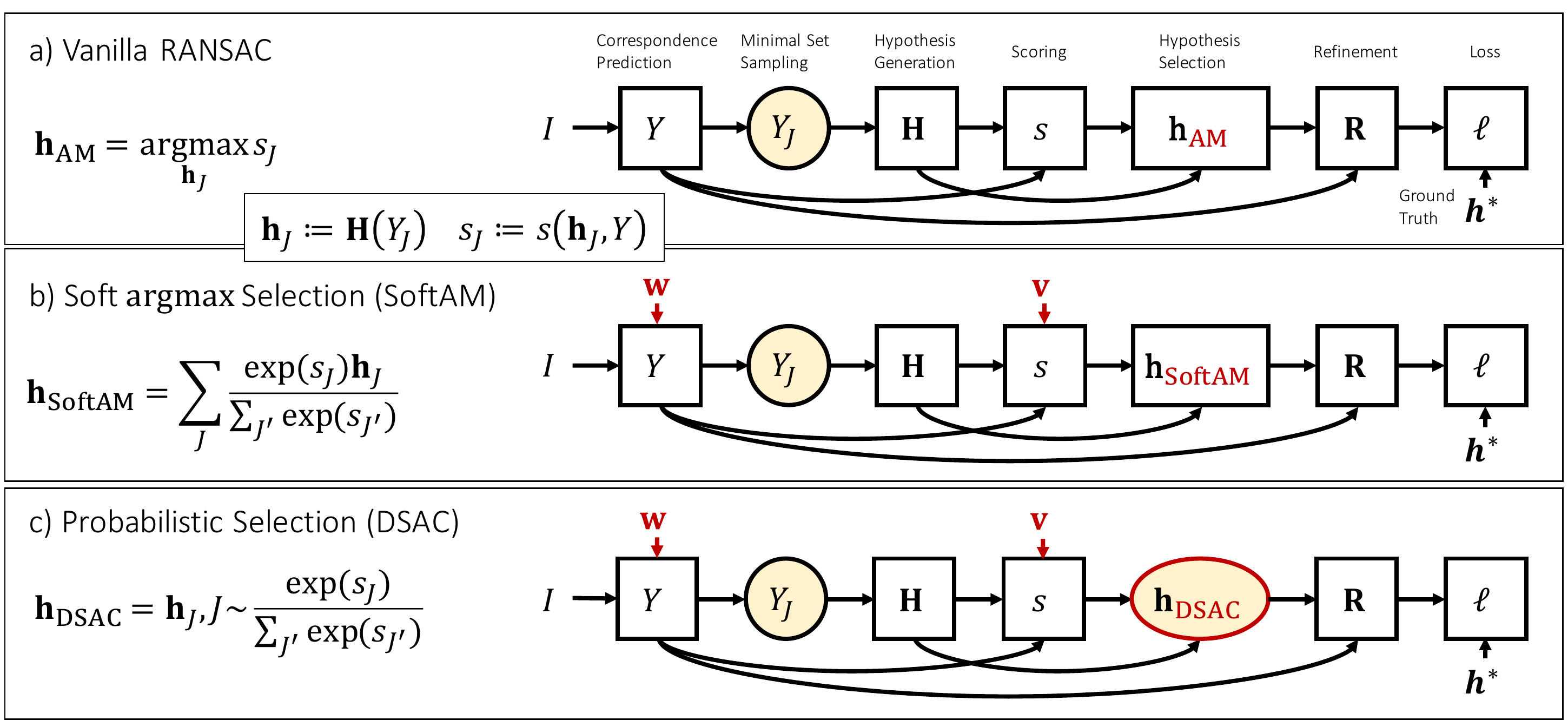}
\caption{{\bf Stochastic Computation Graphs \cite{compgraph2015}.} A graphical representation of three RANSAC variants investigated in this work. The variants differ in the way they select the final model hypothesis: {\bf a)} non-differentiable, vanilla RANSAC with hard, deterministic $\argmax$ selection; {\bf b)} differentiable RANSAC with deterministic, soft $\argmax$ selection; {\bf c)} differentiable RANSAC with hard, probabilistic selection (named DSAC). Nodes shown as boxes represent deterministic functions, while circular nodes with yellow background represent probabilistic functions. Arrows indicate dependency in computation. All differences between a), b) and c) are marked in red.}
\label{fig:stochgraph}
\vspace{-0.5cm}
\end{figure*}

\subsection{Background}
\label{sec:meth:am}

As a preface to explaining our method, we first briefly review the standard RANSAC algorithm for model fitting, and how it can be applied to the camera localization problem using discriminative scene coordinate regression.

Many problems in computer vision involve fitting a model to a set of data points, which in practice usually include outliers due to sensor noise and other factors. The RANSAC algorithm was specifically designed to be able to fit models robustly in the presence of noise \cite{ransac1981}. Dozens of variations of RANSAC exist \cite{mlesac2000,Chum2003,preemptive2003, prosac2005}. We consider a general, basic variant here but the new principles presented in this work can be applied to many RANSAC variants, such as to locally-refined preemptive RANSAC \cite{shotton13scorf}. 

A basic RANSAC implementation consists of four steps: (i) generate a set of model hypotheses by sampling minimal subsets of the data; (ii) score hypotheses based on some measure of consensus, \eg by counting inliers; (iii) select the best scoring hypothesis; (iv) refine the selected hypothesis using additional data points, \eg the full set of inliers.  Step (iv) is optional, though in practice important for high accuracy.

We introduce our notation below using the example application of camera localization. We consider an RGB image $I$ consisting of pixels indexed by $i$. We wish to estimate the parameters $\opt{\mdl}$ of a model that explains $I$. In the camera localization problem this is the 6D camera pose, \ie the 3D rotation and 3D translation of the camera relative to the scene's coordinate frame. Following \cite{shotton13scorf}, we do not fit model $\opt{\mdl}$ directly to image data $I$, but instead make use of intermediate, noisy 2D-3D correspondences predicted for each pixel: $\crds(I) = \left\{ \crd(I, i) | \forall i \right\}$, where $\crd(I, i)$ is the `scene coordinate' of pixel $i$, \ie a discriminative prediction for where the point imaged at pixel $i$ lives in the 3D scene coordinate frame. We will use $\crd_i$ as shorthand for $\crd(I, i)$. $\crds(I)$ denotes the complete set of scene coordinate predictions for image $I$, and we write $\crds$ for $\crds(I)$. To estimate $\opt{\mdl}$ from $\crds$ we apply RANSAC as follows:
\begin{compactenum}

\item \textbf{Generate a pool of hypotheses.} Each hypothesis is generated from a subset of correspondences. This subset contains the minimal number of correspondences to compute a unique solution. We call this a minimal set $\crds_J$ with correspondence indices $J = \{j_1, ..., j_n\}$, where $n$ is the minimal set size. To create the set, we uniformly sample $n$ correspondence indices: $j_m \in [1, \dots, |\crds|]$ to get $\crds_J := \{\crd_{j_1}, ..., \crd_{j_n}\}$.  We assume a function $\pnp$ which generates a model hypothesis as $\mdl_J = \pnp(\crds_J)$ from the minimal set $\crds_J$. In our application, $\pnp$ is the perspective-n-point (PNP) algorithm \cite{gao2003complete}, and $n = 4$.

\item \textbf{Score hypotheses.} Scalar function $s(\mdl_J, \crds)$ measures the consensus / quality of hypothesis $\mdl_J$, \eg by counting inlier correspondences. To define an inlier in our application, we first define the reprojection error of scene coordinate $\crd_i$:
\begin{equation}
\label{eq:ransac:reprerr}
e_i = \|\mathbf{p}_i - C \mdl_J \crd_i\|,
\end{equation}
where $\mathbf{p}_i$ is the 2D location of pixel $i$ and $C$ is the camera projection matrix. We call $\crd_i$ an inlier if $e_i < \tau$, where $\tau$ is the inlier threshold. In this work, instead of counting inliers, we to aim to learn $s(\mdl_J, \crds)$ to directly regress the hypothesis score from reprojection errors $e_i$, as we will explain shortly. 

\item \textbf{Select best hypothesis.} We take 
\begin{equation}
\label{eq:ransac:am}
\am{\mdl} = \argmax_{\mdl_J} s(\mdl_J, \crds) ~.
\end{equation}

\item \textbf{Refine hypothesis.}~$\am{\mdl}$ is refined using function $\refine(\am{\mdl}, \crds)$. Refinement may use all correspondences $\crds$. A common approach is to select a set of inliers from $\crds$ and recalculate function $\pnp$ on this set. The refined pose is the output of the algorithm $\am{\opt{\mdl}} = \refine(\am{\mdl}, \crds)$.
\end{compactenum}

\subsection{Learning in a RANSAC Pipeline}

The system of Shotton \etal \cite{shotton13scorf} had a single learned component, namely the regression forest that made the predictions $\crd(I, i)$.  Krull \etal \cite{krull2015analysis} extended the approach to also learn the scoring function $s(\mdl_J, \crds)$ as a generalization of the simpler inlier counting scheme of \cite{shotton13scorf}. However, these have thus far been learned separately.

Our work instead aims to learn both, the scene coordinate predictions and the scoring function, and to do so jointly in an end-to-end fashion within a RANSAC framework. Making the parameterizations explicit, we have $\crd(I, i; \paramw)$ and $s(\mdl_J, \crds; \paramv)$. We aim to learn parameters $\paramw$ and $\paramv$, where $\paramw$ affects the quality of poses that we generate, and $\paramv$ affects the selection process which should choose a good hypothesis. We write $Y^{\paramw}$ to reflect that scene coordinate predictions depend on parameters $\paramw$. Similarly, we write $\am{\mdl}^{\paramw, \paramv}$ to reflect that the chosen hypothesis depends on $\paramw$ and $\paramv$. 

We would like to find parameters $\paramw$ and $\paramv$ such that the loss $\loss$ of the final, refined hypotheses over a training set of images $\mathcal{I}$ is minimized, \ie
\begin{equation}
\label{eq:meth:ransac:goal}
\opt{\paramw}, \opt{\paramv} = \argmin_{\paramw, \paramv} \sum_{I \in \mathcal{I}} \loss(\refine(\am{\mdl}^{\paramw, \paramv}, \crds^{\paramw}), \gt{\mdl}),
\end{equation}
where $\gt{\mdl}$ are ground truth model parameters for $I$. To allow end-to-end learning, we need to differentiate \wrt $\paramw$ and $\paramv$. We assume a differentiable loss $\loss$ and differentiable refinement $\refine$.

One might consider differentiating $\am{\mdl}^{\paramw, \paramv}$ \wrt to $\paramw$ via the minimal set $\crds_{J}$ of the single selected hypothesis of  Eq.~\ref{eq:ransac:am}. But learning a RANSAC pipeline in this fashion fails because the selection process itself depends on $\paramw$ and $\paramv$, which is not represented in the gradients of the selected hypothesis.\footnote{We observed in early experiments that the training loss immediately increases without recovering.} Parameters $\paramv$ influence the selection directly via the scoring function $s(\mdl, \crds; \paramv)$, and parameters $\paramw$ influence the quality of competing hypotheses $\mdl$, though neither influence the initial uniform sampling of minimal sets $\crds_J$.

We next present two approaches to learn parameters $\paramw$ and $\paramv$ -- soft $\argmax$ selection (Sec.~\ref{sec:meth:softam}) and probabilistic selection (Sec.~\ref{sec:meth:dsac}) -- that do model the dependency of the selection process on the parameters.

\subsubsection{Soft $\argmax$ Selection (SoftAM)}
\label{sec:meth:softam}

To solve the problem of non-differentiability, one can relax the $\argmax$ operator of Eq.~\ref{eq:ransac:am} and substitute it for a soft $\argmax$ operator \cite{Chapelle2010}. The soft $\argmax$ turns the hypothesis selection into a weighted average of hypotheses:
\begin{equation}
\label{eq:ransac:samax:avg}
\sam{\mdl}^{\paramw, \paramv} = \sum_J P(J|\paramv,\paramw) \mdl_J^\paramw
\end{equation}
which averages over candidate hypotheses $\mdl_J^\paramw$ with
\begin{equation}
\label{eq:ransac:samax}
P(J|\paramv,\paramw) = \frac{\exp(s(\mdl_J^\paramw, \crds^\paramw; \paramv))}{\sum_{J'} \exp(s(\mdl_{J'}^\paramw \crds^\paramw; \paramv))} ~.
\end{equation}
In this variant, scoring function $s(\mdl_J^\paramw, \crds^\paramw; \paramv)$ has to predict weights that lead to a robust average of hypotheses (\ie model parameters). This means that model parameters corrupted by outliers should receive sufficiently small weights, such that they do not affect the accuracy of $\sam{\mdl}^{\paramw, \paramv}$.

Substituting $\am{\mdl}^{\paramw, \paramv}$ for $\sam{\mdl}^{\paramw, \paramv}$ in Eq.~\ref{eq:meth:ransac:goal} allows us to calculate gradients to learn parameters $\paramw$ and $\paramv$. We refer the reader to the appendix for details.

By utilizing the soft $\argmax$ operator, we diverge from the RANSAC principle of making one hard decision for a hypothesis. Soft $\argmax$ hypothesis selection bears similarity with an independent strain within the field of robust optimization, namely robust averaging, see \eg the work of Hartley \etal \cite{rotavg2011}. While we explore soft $\argmax$ selection in the experimental evaluation, we introduce an alternative in the next section, that preserves the hard hypothesis selection, and is empirically superior for our task.

\subsubsection{Probabilistic Selection (DSAC)}
\label{sec:meth:dsac}

We substitute the deterministic selection of the highest scoring model hypothesis in Eq.~\ref{eq:ransac:am} by a probabilistic selection, \ie we chose a hypothesis probabilistically according to:
\begin{equation}
\label{eq:ransac:smpl}
\smpl{\mdl}^{\paramw, \paramv} = \mdl_{J}^\paramw \text{, with } J \sim P(J|\paramv,\paramw),
\end{equation} 
where $P(J|\paramv,\paramw)$ is the softmax distribution of scores predicted by $s(\mdl_J^\paramw, \crds^\paramw; \paramv)$ (see Eq.~\ref{eq:ransac:samax}). 

The inspiration for this approach comes from policy gradient approaches in reinforcement learning that involve the minimization of a loss function defined over a stochastic process \cite{compgraph2015}. Similarly, we are able to learn parameters $\paramw$ and $\paramv$ that minimize the expectation of loss of the stochastic process defined in Eq.~\ref{eq:ransac:smpl}:
\begin{equation}
\opt{\paramw}, \opt{\paramv} = \argmin_{\paramw, \paramv} \sum_{I \in \mathcal{I}} \expectation{J \sim P(J|\paramv,\paramw)}{\loss(\refine(\mdl_J^\paramw, \crds^\paramw))}.
\end{equation}
As shown in \cite{compgraph2015}, we can calculate the derivative \wrt parameters $\paramw$ as follows (similarly for parameters $\paramv$):
\begin{multline}
\label{eq:ransac:smpl:derv}
\derv{\paramw}\expectation{J \sim P(J|\paramv,\paramw)}{\loss(\cdot)} = \\
\expectation{J \sim P(J|\paramv,\paramw)}{\loss(\cdot) \derv{\paramw} \log P(J|\paramv,\paramw) + \derv{\paramw}\loss(\cdot)}, 
\end{multline}
\ie the derivative of the expectation is an expectation over derivatives of the loss and the $\log$ probabilities of model hypotheses. We include further steps of the derivation of Eq.~\ref{eq:ransac:smpl:derv} in the appendix.

We call this method of differentiating RANSAC, that preserves hard hypothesis selection, DSAC -- Differentiable SAmple Consensus. See Fig.~\ref{fig:stochgraph} for a schematic view of DSAC in comparison to the RANSAC variants introduced at the beginning of this section. While learning parameters with the vanilla RANSAC is not possible, as mentioned before, both new variants (SoftAM and DSAC) are sensible options which we evaluate in the experimental section.

\section{Differentiable Camera Localization}
\label{sec:diffloc}

\begin{figure*}[h!t]
\includegraphics[scale=0.65]{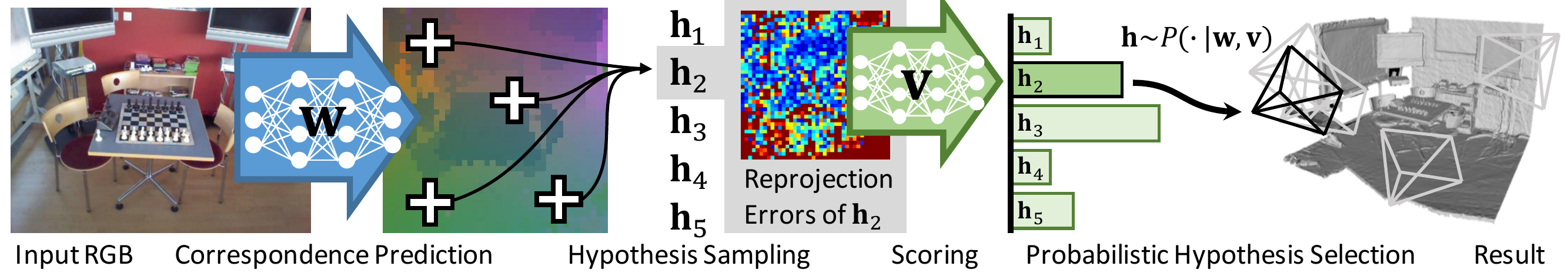}
\caption{{\bf Differentiable Camera Localization Pipeline.}  Given an RGB image, we let a CNN with parameters $\paramw$ predict 2D-3D correspondences, so called scene coordinates \cite{shotton13scorf}. From these, we sample minimal sets of four scene coordinates and create a pool of hypotheses $\mdl$. For each hypothesis, we create an image of reprojection errors which is scored by a second CNN with parameters $\paramv$. We select a hypothesis probabilistically according to the score distribution. The selected pose is also refined.}
\label{fig:exp:sys}
\vspace{-0.5cm}
\end{figure*}

We demonstrate the principles for differentiating RANSAC for the task of one-shot camera localization from an RGB image. Our pipeline is inspired by the state-of-the-art pipeline of Brachmann \etal \cite{brachmann2016}, which is an extension of the original SCoRF pipeline \cite{shotton13scorf} from RGB-D to RGB images. Brachmann \etal use an auto-context random forest to predict multi-modal scene coordinate distributions per image patch. After that, minimal sets of four scene coordinates are randomly sampled and the PNP algorithm \cite{gao2003complete} is applied to create a pool of camera pose hypotheses. A preemptive RANSAC schema iteratively refines, re-scores and rejects hypotheses until only one remains. The preemptive RANSAC scores hypotheses by counting inlier scene coordinates, \ie scene coordinates $\crd_i$ for which reprojection error $e_i < \tau$. In a last step, the final, remaining hypothesis is further optimized using the uncertainty of the scene coordinate distributions. 

Our pipeline differs from Brachmann \etal \cite{brachmann2016} in the following aspects: 
\begin{compactitem}

\item Instead of a random forest, we use a CNN (called `Coordinate CNN' below) to predict scene coordinates. For each 42x42 pixel image patch, it predicts a scene coordinate point estimate. We use a VGG style architecture with 13 layers and 33M parameters. To reduce test time we process only 40x40 patches per image.

\item We score hypotheses using a second CNN (called `Score CNN' below). We took inspiration from the work of Krull \etal \cite{krull2015analysis} for the task of object pose estimation. Instead of learning a CNN to compare rendered and observed images as in \cite{krull2015analysis}, our Score CNN predicts hypothesis consensus based on reprojection errors. For each of the 40x40 scene coordinate predictions $\crd_i$ we calculate the reprojection error $e_i$ for hypothesis $\mdl_J$ (see Eq.~\ref{eq:ransac:reprerr}). This results in a 40x40 reprojection error image, which we feed into the Score CNN, a VGG style architecture with 13 layers and 6M parameters.

\item  Instead of the preemptive RANSAC schema, we score hypotheses only once and select the final pose, either by applying the soft $\argmax$ operator (SoftAM), or by probabilistic selection according to the softmaxed scores (DSAC). 

\item Only the final pose is refined. We choose inlier object coordinate predictions (at most 100), \ie scene coordinates $\crd_i$ with reprojection error $e_i < \tau$, and solve PNP \cite{lepetit2009epnp} again using this set. This is iterated multiple times. Since the Coordinate CNN predicts only point estimates we do no further pose optimization using uncertainty.

\end{compactitem}

See Fig. \ref{fig:exp:sys} for an overview of our pipeline. Where applicable we use the parameter values reported by Brachmann \etal in \cite{brachmann2016}, \eg sampling 256 hypotheses, using 8 refinement steps and an inlier threshold of $\tau = 10$px. 

\section{Experiments}
\label{sec:exp}

For comparability to other methods, we show results on the widely used 7-Scenes dataset \cite{shotton13scorf}. The dataset consists of RGB-D images of 7 indoor environments where each frame is annotated with its 6D camera pose. A 3D model of each scene is also available. The data of each scene is comprised of multiple sequences (= independent camera paths) which are assigned either to test or training. The number of images per scene ranges from 1k to 7k for training resp. test. We omit the depth channels and estimate poses using RGB images only. See the appendix for a discussion of the difficulty of the 7-Scenes dataset.

We measure accuracy by the percentage of images for which the camera pose error is below $5^\circ$ and $5$cm (see Appendix \ref{app:error} for a comment on the calculation of this error). For training, we use the following differentiable loss which is closely correlated with the task loss:
\begin{equation}
\label{eq:exp:diffloss}
\loss_{\text{pose}}(\mdl, \gt{\mdl}) = \max(\measuredangle(\boldsymbol{\theta}, \gt{\boldsymbol{\theta}}), \| \mathbf{t} - \gt{\mathbf{t}} \|),
\end{equation}
where $\mdl = (\boldsymbol{\theta}, \mathbf{t})^{-1}$, $\boldsymbol{\theta}$ denotes the axis-angle representation of the camera rotation, and $\mathbf{t}$ is the camera translation. We measure angle $\measuredangle(\boldsymbol{\theta}, \gt{\boldsymbol{\theta}})$ between estimated and ground truth rotation in degree, and distance $\| \mathbf{t}- \gt{\mathbf{t}} \|$ between estimated and ground truth translation in cm.

Since the dataset does not include a designated validation set, we separated multiple blocks of $100$ consecutive frames from the training data to be used as validation data (in total $10\%$ per scene). We fixed all learning parameters on the validation set (\eg learning rate and total amount of parameter updates). Once all hyper parameters are fixed, we re-train on the full training set.

\subsection{Componentwise Training}
\label{sec:exp:componentwise}

Our pipeline contains two trainable components, namely the Coordinate CNN and the Score CNN. First, we explain how to train both components using surrogate losses, \ie train them not in an end-to-end fashion but separately. End-to-end training using differentiable RANSAC will be discussed in Sec.~\ref{sec:exp:endtoend}.

\noindent{\bf Scene Coordinate Regression.} Similar to Brachmann \etal \cite{brachmann2016}, we use the depth information of training images to generate scene coordinate ground truth. Alternatively, this ground truth can also be rendered using the available 3D models. We train the Coordinate CNN using the following surrogate loss: $\loss_{\text{coord}}(\crd, \gt{\crd}) = \| \crd - \gt{\crd} \|$, where $\crd$ is the scene coordinate prediction and $\gt{\crd}$ is ground truth. We also experimented with other losses including $L_2$ (squared distance), Huber \cite{huber1964} and Tukey \cite{beaton1974} which consistently performed worse on the validation set.

We trained with mini batches of 64 randomly sampled training patches. We used the Adam \cite{adam2014} optimizer with a learning rate of $10^{-4}$. We cut the learning rate in half after each 50k updates, and train for a total of 300k updates.

\noindent{\bf Score Regression.} We synthetically created data to train the Score CNN in the following way. By adding noise to the ground truth pose of training images, we generated poses above and below the pose error threshold of $5^\circ$ and $5$cm. Using the scene coordinate predictions of the trained Coordinate CNN, we compute reprojection error images of these poses. Poses with a large pose error \wrt the ground truth pose will lead to large reprojection errors, and we want the Score CNN to predict a small score. Poses close to ground truth will lead to small reprojection errors, and we want the Score CNN to predict a high score. More formally, the pose error $\loss_{\text{pose}}(\mdl, \gt{\mdl})$ of a hypothesis $\mdl$ should be negatively correlated with the score prediction $s(\mdl, \crds; \paramv)$. Thus, we train the Score CNN to minimize the following loss: $\loss_{\text{score}}(s, \gt{s}) = |s -  \gt{s}|$, where:
$\gt{s} = - \beta \loss_{\text{pose}}(\mdl, \gt{\mdl})$. Parameter $\beta$ controls the broadness of the score distribution after applying softmax. We use this distribution for weights in SoftAM (see Eq.~\ref{eq:ransac:samax}) and to sample a hypothesis in DSAC (see Eq.~\ref{eq:ransac:smpl}). A value of $\beta=10$ gave reasonable distributions on the validation set, \ie poses close to ground truth had a high probability to be selected, and poses far away from ground truth had a low probability to be selected. 

We trained the Score CNN with a batch size of 64 reprojection error images of randomly generated poses. We used Adam \cite{adam2014} for optimization with a learning rate of $10^{-4}$. We train for a total of 2k updates.

\begin{table*}[h!t]
\centering
\caption{Accuracy measured as the percentage of test images where the pose error is below 5cm and 5$^\circ$. \emph{Complete} denotes the combined set of frames (17000) of all scenes. Numbers in \textcolor{green}{green} denote improved accuracy after end-to-end training for SoftAM resp. DSAC compared to componentwise training. Similarly, \textcolor{red}{red} numbers denote decreased accuracy. {\bf Bold} numbers indicate the best result for each scene.}
\label{exp:res:acc}
\begin{tabular}{lcc|c|c|c|c|c|}
\cline{2-8}
\multicolumn{1}{l|}{}         & \multicolumn{1}{c|}{Sparse}          & Brachmann     & \multicolumn{3}{c|}{Ours: Trained Componentwise} & \multicolumn{2}{c|}{Ours: Trained End-To-End} \\ \cline{4-8} 
\multicolumn{1}{l|}{}         & \multicolumn{1}{c|}{Features \cite{shotton13scorf}} & \etal \cite{brachmann2016} & RANSAC     & SoftAM    & DSAC      & SoftAM         & DSAC          \\ \hline
\multicolumn{1}{|l|}{Chess}   & \multicolumn{1}{c|}{70.7\%}  & \textbf{94.9\%}        & \textbf{94.9\%}     & 94.8\%     & 94.7\%    & 94.2\% \textcolor{red}{-0.6\%}   & 94.6\% \textcolor{red}{-0.1\%}       \\
\multicolumn{1}{|l|}{Fire}    & \multicolumn{1}{c|}{49.9\%}  & 73.5\%        & 75.1\%     & 75.6\%     & 75.3\%    & \textbf{76.9\%} \textcolor{green}{+1.3\%} & 74.3\% \textcolor{red}{-1.0\%}       \\
\multicolumn{1}{|l|}{Heads}   & \multicolumn{1}{c|}{67.6\%}  & 48.1\%        & 72.5\%     & \textbf{74.5\%}     & 71.9\%    & 74.0\% \textcolor{red}{-0.5\%}   & 71.7\% \textcolor{red}{-0.2\%}       \\
\multicolumn{1}{|l|}{Office}  & \multicolumn{1}{c|}{36.6\%}  & 53.2\%        & 70.4\%     & \textbf{71.3\%}     & 69.2\%    & 56.6\% \textcolor{red}{-14.7\%}  & 71.2\% \textcolor{green}{+2.0\%}       \\
\multicolumn{1}{|l|}{Pumpkin} & \multicolumn{1}{c|}{21.3\%}  & \textbf{54.5\%}        & 50.7\%     & 50.6\%     & 50.3\%    & 51.9\% \textcolor{green}{+1.3\%} & 53.6\% \textcolor{green}{+3.3\%}       \\
\multicolumn{1}{|l|}{Kitchen} & \multicolumn{1}{c|}{29.8\%}  & 42.2\%        & 47.1\%     & 47.8\%     & 46.2\%    & 46.2\% \textcolor{red}{-1.6\%}   & \textbf{51.2\%} \textcolor{green}{+5.0\%}       \\
\multicolumn{1}{|l|}{Stairs}  & \multicolumn{1}{c|}{9.2\%}   & \textbf{20.1\%}        & 6.2\%      & 6.5\%      & 5.3\%     & 5.5\% \textcolor{red}{-1.0\%}    & 4.5\% \textcolor{red}{-0.8\%}       \\ \hline
\multicolumn{1}{|l|}{Average} & \multicolumn{1}{c|}{40.7\%}  & 55.2\%        & 59.5\%     & \textbf{60.1\%}     & 59.0\%    & 57.9\% \textcolor{red}{-2.2\%}   & \textbf{60.1\%} \textcolor{green}{+1.1\%}       \\ \hline
\multicolumn{1}{|l|}{Complete} & \multicolumn{1}{c|}{38.6\%} & 55.2\%        & 61.0\%     & 61.6\%     & 60.3\%    & 57.8\% \textcolor{red}{-3.8\%}   & \textbf{62.5\%} \textcolor{green}{+2.2\%}       \\ \hline
\end{tabular}
\end{table*}

\begin{table}[h!]
\centering
\caption{{\bf Median pose errors} of the complete 7-Scenes dataset (17000 frames). Most accurate results marked {\bf bold}.}
\label{exp:res:err}
\begin{tabular}{c|l|l|}
\cline{2-3}
\multicolumn{1}{l|}{}                                                                                   & \begin{tabular}[c]{@{}l@{}}Brachmann \\ \etal \cite{brachmann2016}\end{tabular} & 4.5cm, 2.0$^\circ$ \\ \hline \hline
\multicolumn{1}{|c|}{\multirow{3}{*}{\begin{tabular}[c]{@{}c@{}}Ours, Trained\\ Componentwise\end{tabular}}} & RANSAC                                                             & 4.0cm, 1.6$^\circ$ \\ \cline{2-3} 
\multicolumn{1}{|c|}{}                                                                                  & SoftAM                                                            & {\bf 3.9cm, 1.6$^\circ$} \\ \cline{2-3} 
\multicolumn{1}{|c|}{}                                                                                  & DSAC                                                               & 4.0cm, 1.6$^\circ$ \\ \hline \hline
\multicolumn{1}{|c|}{\multirow{2}{*}{\begin{tabular}[c]{@{}c@{}}Ours, Trained \\ End-To-End\end{tabular}}}    & SoftAM                                                            & 4.0cm, 1.6$^\circ$ \\ \cline{2-3} 
\multicolumn{1}{|c|}{}                                                                                  & DSAC                                                               & {\bf 3.9cm, 1.6$^\circ$} \\ \hline
\end{tabular}
\end{table}

\noindent{\bf Results.} We report the accuracy of our pipeline, trained componentwise, in Table \ref{exp:res:acc}. We present the accuracy per scene and the average over scenes. Since scenes with few test frames like \emph{Stairs} and \emph{Heads} are overrepresented in the average, we additionally show accuracy on the dataset as a whole (denoted \emph{Complete}, \ie 17000 test frames).

We distinguish between \emph{RANSAC}, \ie non-differentiable $\argmax$ hypothesis selection, \emph{SoftAM}, \ie differentiable soft $\argmax$ hypothesis selection and \emph{DSAC}, \ie differentiable probabilistic hypothesis selection.

As can be seen in Table \ref{exp:res:acc}, RANSAC, SoftAM and DSAC achieve very similar results when trained componentwise. The probabilistic hypothesis selection of DSAC results in a slightly reduced accuracy of -0.7\% on the complete dataset, compared to RANSAC. 

We compare our pipeline to the sparse features baseline presented in \cite{shotton13scorf} and the pipeline of Brachmann \etal \cite{brachmann2016}, which is state-of-the-art on this dataset at the moment. All variants of our pipeline surpass, on average, the accuracy of both competitors. Note, conceptually the main advantage over Brachmann \etal \cite{brachmann2016} is the new scoring CNN. We also measured the median pose error of all frames in the dataset, see Table \ref{exp:res:err}. Compared to Brachmann \etal \cite{brachmann2016} we are able to decrease both rotational and translational error. PoseNet \cite{kendall2015convolutional} states median translational errors of around 40cm per scene, so it cannot compete in terms of accuracy.

\subsection{End-to-End Training}
\label{sec:exp:endtoend}

In order to facilitate end-to-end learning as described in Sec.~\ref{sec:method}, some parts of the pipeline need to be differentiable which might not be immediately obvious. We already introduced the differentiable loss $\loss_{\text{pose}}$. Furthermore, we need to derive the model function $\pnp(\crds_J)$ and refinement $\refine$ \wrt learnable parameters.

In our application, $\pnp(\crds_J)$ is the PNP algorithm. Off-the-shelf implementations (\eg \cite{gao2003complete, lepetit2009epnp}) are fast enough for calculating the derivatives via central differences. 

Refinement $\refine$ involves determining inlier sets and resolving PNP in multiple iterations. This procedure in non-differentiable because of the hard inlier selection procedure. However, because the number of inliers is large (100 in our case), refined poses tend to vary smoothly with changes to the input scene coordinates. Hence, we treat the refinement procedure as a black box, and calculate derivatives via central differences, as well. For stability, we stop refinement early, in case less than 50 inliers have been found. Because of the large number of inputs and to keep central differences tractable, we subsample the scene coordinates for which gradients are calculated (we use 1\%), and correct the gradient magnitude accordingly ($\times 100$). 

Similar to \eg \cite{lift2016} or \cite{kendall2015convolutional}, we found it important to have a good initialization when learning end-to-end. Learning from scratch quickly reached a local minimum. Hence, we initialize the Coordinate CNN and the Score CNN with componentwise training, see Sec.~\ref{sec:exp:componentwise}.

We found the same set of training hyperparameters to work well for the validation set for both, SoftAM and DSAC. We use a fixed learning rate of $10^{-5}$ for the Coordinate CNN, and a fixed learning rate of $10^{-7}$ for the Score CNN. Our end-to-end pipeline contains substantial stochasticity because of the sampling of minimal sets $\crds_J$. Instead of the Adam procedure, which was unstable, we use stochastic gradient descent with momentum \cite{backprob1988} of 0.9, and we clamp all gradients to the range of -0.1 to 0.1, before passing them to the Score CNN or the Coordinate CNN. We train for 5k updates.

\noindent{\bf Results.} See Table \ref{exp:res:acc} for results of both strategies. Compared to the initialization (trained componentwise), we observe a significant improvement for DSAC (+2.2\% on the complete dataset, standard error of the mean $\pm0.4\%$). DSAC improves some scenes considerably, with strongest effects for \emph{Pumpkin} (+3.3\%) and \emph{Kitchen} (+5.0\%). SoftAM significantly decreases accuracy compared to the componentwise initialization (-3.8\% on the complete dataset). SoftAM overfits severely on the \emph{Office} scene (-14.7\%) and decreases accuracy for most other scenes. 

The pipeline learned end-to-end with DSAC improves on the results of Brachmann \etal \cite{brachmann2016} by 4.9\% (scene average) resp.~7.3\% (complete set). DSAC also improves the median pose error, see Table~\ref{exp:res:err}.

\subsection{Insights and Detailed Studies}

\begin{figure}[h!]

   \includegraphics[width=1.0\linewidth]{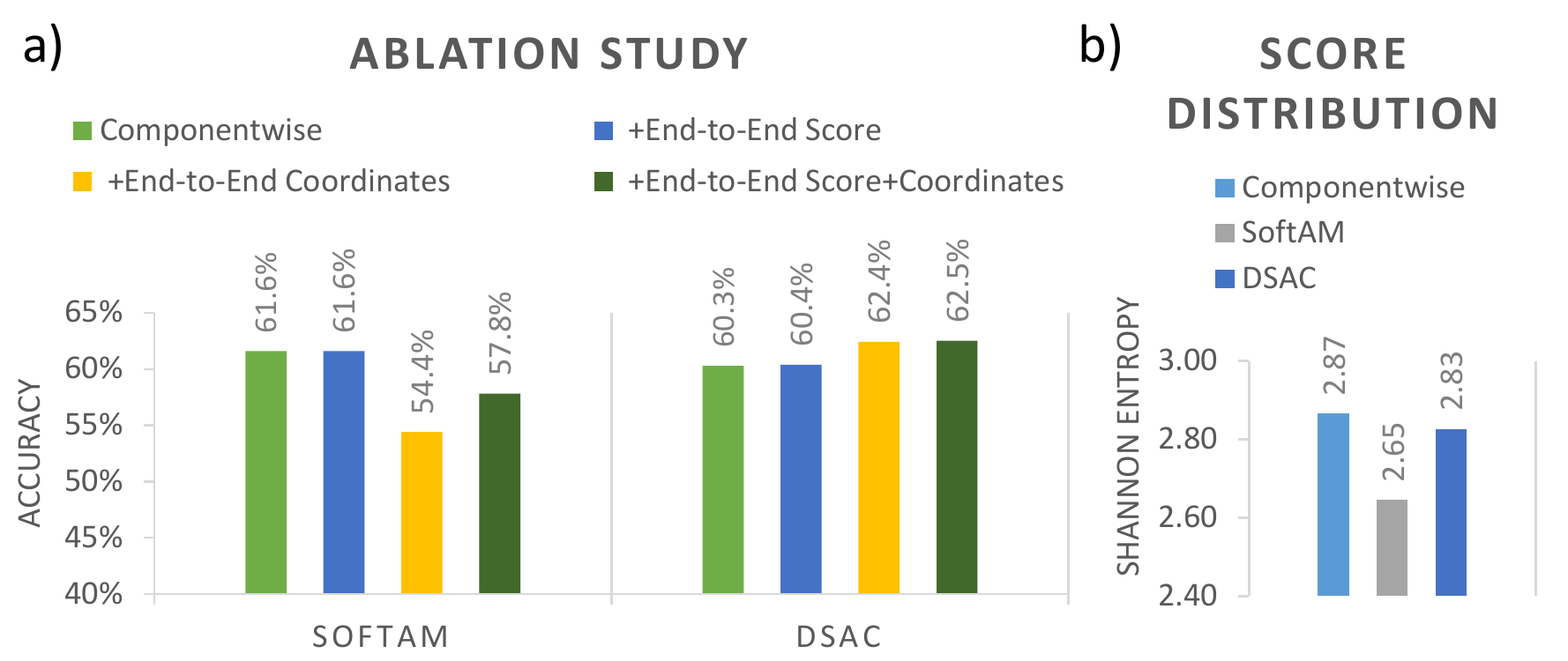}

   \caption{{\bf (a)} Effect of end-to-end learning on pose accuracy \wrt individual components. {\bf (b)} Effect of end-to-end training on the average entropy of the score distribution. Set text for details.}
\label{fig:exp:train}
\end{figure}

\noindent{\bf Ablation Study.} We study the effect of learning the Score CNN and the Coordinate CNN in an end-to-end fashion,  individually. We use componentwise training as initialization for both CNNs. See Fig.~\ref{fig:exp:train} a) for results on the complete set. For DSAC, training both components in an end-to-end fashion is important for best accuracy. For SoftAM, we see that the bad results on this scene are not due to overfitting on the Score CNN, but its way of learning the Coordinate CNN.

\begin{figure}[h!t]
  \includegraphics[width=1.0\linewidth]{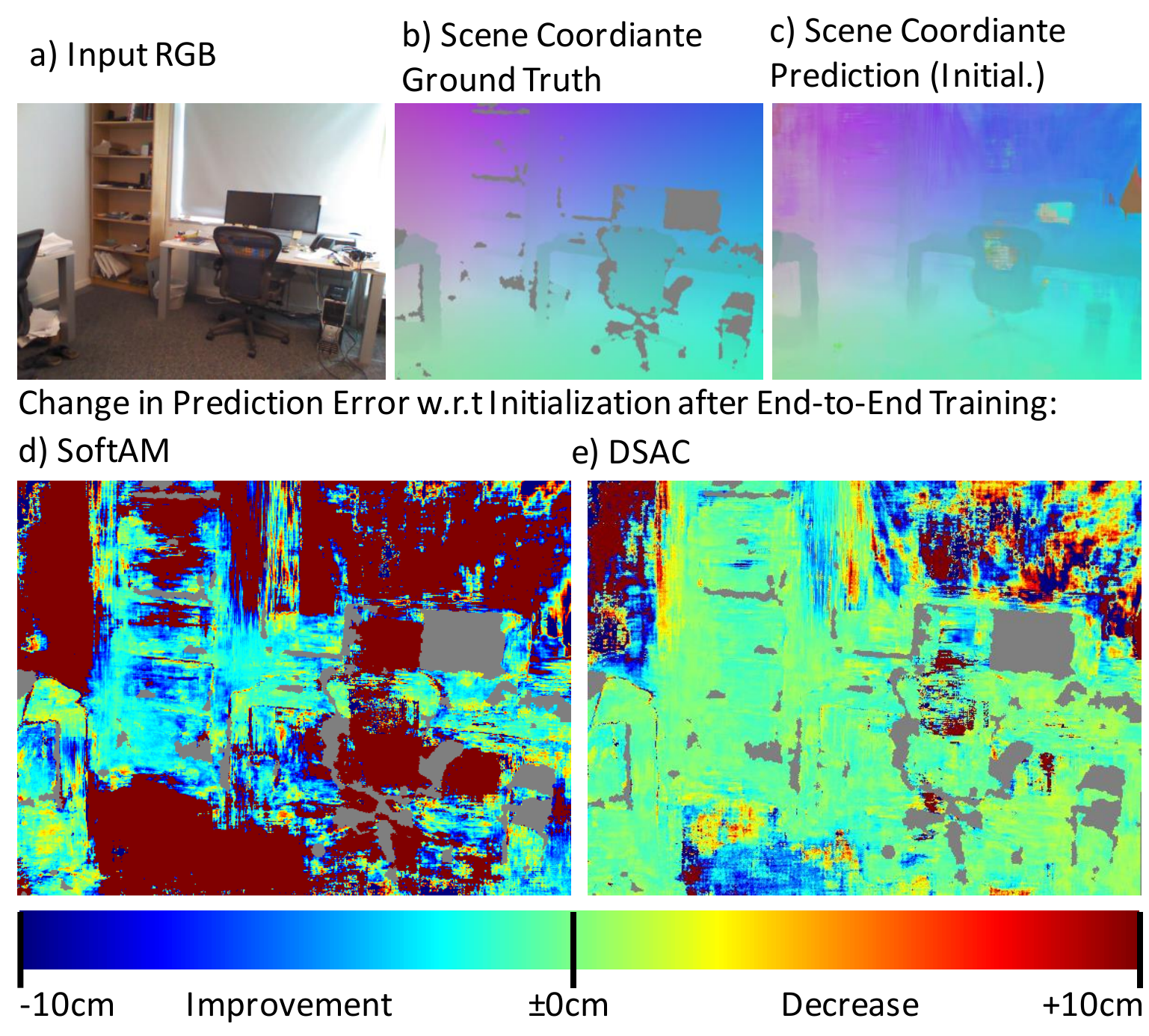}
   \caption{{\bf Prediction quality.} We analyze scene coordinate prediction quality on an \emph{Office} test image {\bf (a)} with ground truth scene coordinates {\bf (b)} (XYZ mapped to RGB). The prediction after componentwise training can be seen in {\bf (c)}. We vizualize the relative change of prediction error \wrt componentwise training in {\bf (d)} for SoftAM, resp.~in {\bf (e)} for DSAC. We observe an aggressive strategy of SoftAM which focuses large improvements on small areas (14\% of predictions improve). DSAC shows small improvements but on large areas (38\% of predictions improve). Note that DSAC achieves superior pose accuracy on this scene.}
\label{fig:exp:change}
\end{figure}

\noindent{\bf Analysis of Scene Coordinate Predictions.} In the componentwise training, the Coordinate CNN learned to minimize the surrogate loss $\loss_{\text{coord}}$, \ie the distance $\|\crd_i - \gt{\crd}_i \|$ of scene coordinate predictions $\crd_i$ \wrt ground truth $\gt{\crd}_i$. In Fig.~\ref{fig:exp:change}, we visualize how the prediction of the Coordinate CNN changes when trained in an end-to-end fashion, \ie to minimize the loss $\loss_{\text{pose}}$. Both end-to-end learning strategies, SoftAM and DSAC, increase the accuracy of scene coordinate predictions in some areas of the scene at the cost of decreasing the accuracy in other areas. We observe very extreme changes for the SoftAM strategy, \ie the increase and decrease in scene coordinate accuracy is large in magnitude, and improvements are focused to small scene areas. The DSAC strategy leads to a much more cautious tradeoff, \ie changes are smaller and widespread. Note that we use identical learning parameters for both strategies. We conclude that SoftAM tends to overfit due to overly aggressive changes in scene coordinate predictions. 

\noindent{\bf Score Distribution Entropy.} See Fig.~\ref{fig:exp:train} b) for an analysis of the effect of end-to-end learning on the average entropy of the softmax score distribution (see Eq.~\ref{eq:ransac:samax}). We observe a reduction in entropy for the SoftAM strategy. The larger the pose error of a hypothesis is, the larger is also its influence on the pose average (see Eq.~\ref{eq:ransac:samax:avg}). SoftAM has to weigh down such poses aggressively for a good average. DSAC can allow for a broader distribution (only a slight decrease in entropy compared to the original RANSAC) because poses which are unlikely to be chosen, do not affect the loss of poses which are likely to be chosen. This is an additional factor in the stability of DSAC.

\noindent{\bf Restoring the $\argmax$ Selection.} After end-to-end training, one may restore the original RANSAC algorithm, \eg selecting hypotheses \wrt scores via $\argmax$. In this case, the average accuracy of DSAC stays at $62.4\%$, while the accuracy of SoftAM decreases to $57.2\%$. 

\noindent{\bf Test Time.} The scene coordinate prediction takes $\sim$0.5s on a Tesla K80 GPU. Pose optimization takes $\sim$1s. The runtime of $\argmax$ hypothesis selection (RANSAC) or probabilistic selection (DSAC) is identical and negligible.

\noindent{\bf Multi-Modality.} Compared to Brachmann \etal \cite{brachmann2016}, our pipeline performs not as well on the \emph{Stairs} scene (see Table \ref{exp:res:acc}). We account this to the fact that the Coordinate CNN predicts only uni-modal point estimates, whereas the random forest of \cite{brachmann2016} predicts multi-modal scene coordinate distributions. The \emph{Stairs} scene contains many repeating structures, so we expect multi-modal predictions to help. We also expect bad performance of the SoftAM strategy in case pose hypothesis distributions are multi-modal, because an average is likely to be a bad representation of either mode. In contrast, DSAC can probabilistically select the correct mode. We conclude that multi-modality in scene coordinate predictions and pose hypothesis distributions is a promising direction for future work.

\section{Conclusion}
\label{sec:con}

We presented two strategies for differentiating the RANSAC algorithm: Using a soft $\argmax$ operator, and probabilistic selection. By experimental evaluation we conclude that probabilistic selection is superior and call this approach DSAC.  We demonstrated the use of DSAC for learning a camera localization pipeline end-to-end. However, DSAC can be deployed in any deep learning pipeline where robust optimization is beneficial, for example learning structure from motion or SLAM end-to-end. 

\paragraph*{Acknowledgements:}
This project has received funding from the European Research Council (ERC) under the European Union’s Horizon 2020 research and innovation programme (grant agreement No 647769). The computations were performed on an HPC Cluster at the Center for Information Services and High Performance Computing (ZIH) at TU Dresden. We thank the Torr Vision Group of the University of Oxford for inspiring discussions.

\appendix

\section{Derivatives}

This appendix contains additional information on the derivative of the task loss function (resp.~the expectation thereof) for the SoftAM and DSAC learning strategies. In the second part of the appendix, we illustrate some  difficulties of camera localization on the 7-Scenes dataset to motivate the usage of a RANSAC schema for this problem. 

\subsection{Soft $\argmax$ Selection (SoftAM)}

To learn our camera localization pipeline in an end-to-end fashion, we have to calculate the derivatives of the task loss function $\loss(\refine(\sam{\mdl}^{\paramw, \paramv}, \crds^{\paramw}), \gt{\mdl})$ \wrt to learnable parameters. In the following, we show the derivative \wrt parameters $\paramw$, but derivation \wrt parameters $\paramv$ works similarly.   
Applying the chain rule and calculating the total derivative of $\refine$, we get:
\vspace{-0.2cm}
\begin{multline}
\label{eq:appendix:lderv}
\derv{\paramw}\loss(\refine(\sam{\mdl}^{\paramw, \paramv}, \crds^{\paramw}), \gt{\mdl}) = \\
\frac{\partial \loss}{\partial \refine} \left( \frac{\partial \refine}{\partial \sam{\mdl}^{\paramw, \paramv}} \frac{\partial \sam{\mdl}^{\paramw, \paramv}}{\partial \paramw} + \frac{\partial \refine}{\partial \crds^\paramw} \frac{\partial \crds^\paramw}{\partial \paramw} \right)
\end{multline}
Since $\sam{\mdl}^{\paramw, \paramv}$ is a weighted average of hypothesis (see Eq.~\ref{eq:ransac:samax:avg}) we can differentiate it as follows:
\vspace{-0.3cm}
\begin{multline}
\derv{\paramw} \sam{\mdl}^{\paramw, \paramv} = \\
\sum_J \left( \left(\derv{\paramw} P(J|\paramv,\paramw) \right) \mdl_J^\paramw + P(J|\paramv,\paramw) \derv{\paramw} \mdl_J^\paramw \right)
\end{multline}
Weights $P(J|\paramv,\paramw)$ follow a softmax distribution of hypothesis scores (see Eq.~\ref{eq:ransac:samax}). 
Hence, we can differentiate as follows:
\vspace{-0.3cm}
\begin{multline}
\derv{\paramw} P(J|\paramv,\paramw) = P(J|\paramv,\paramw) \\
\left( \derv{\paramw} s(\mdl_J^\paramw, \paramv) 
- \expectation{J' \sim P(J'|\paramv,\paramw)}{\derv{\paramw}s(\mdl_{J'}^\paramw, \paramv)} \right)
\end{multline}

\subsection{Probabilistic Selection (DSAC)}

Using the DSAC strategy, we learn our camera localization pipeline by minimizing the expectation of the task loss function: 
\begin{multline}
\derv{\paramw}\expectation{J \sim P(J|\paramv,\paramw)}{\loss(\cdot)} = \\
\expectation{J \sim P(J|\paramv,\paramw)}{\loss(\cdot) \derv{\paramw} \log P(J|\paramv,\paramw) + \derv{\paramw}\loss(\cdot)},
\end{multline}
where we use $\loss(\cdot)$ as a stand-in for $\loss(\refine(\mdl_J^{\paramw, \paramv}, \crds^{\paramw}), \gt{\mdl})$. We differentiate $\derv{\paramw}\loss(\cdot)$ following Eq.~\ref{eq:appendix:lderv}, and log probabilities $\log P(J|\paramv,\paramw)$ as:
\begin{multline}
\derv{\paramw} \log P(J|\paramv,\paramw) = \\
\derv{\paramw} s(\mdl_J^\paramw, \paramv) - \expectation{J' \sim p(J'|\paramv,\paramw)}{\derv{\paramw} s(\mdl_{J'}^\paramw, \paramv)}.
\end{multline}

\section{Difficulty of the 7-Scenes Dataset}

\begin{figure}[h!t]

   \includegraphics[width=1.0\linewidth]{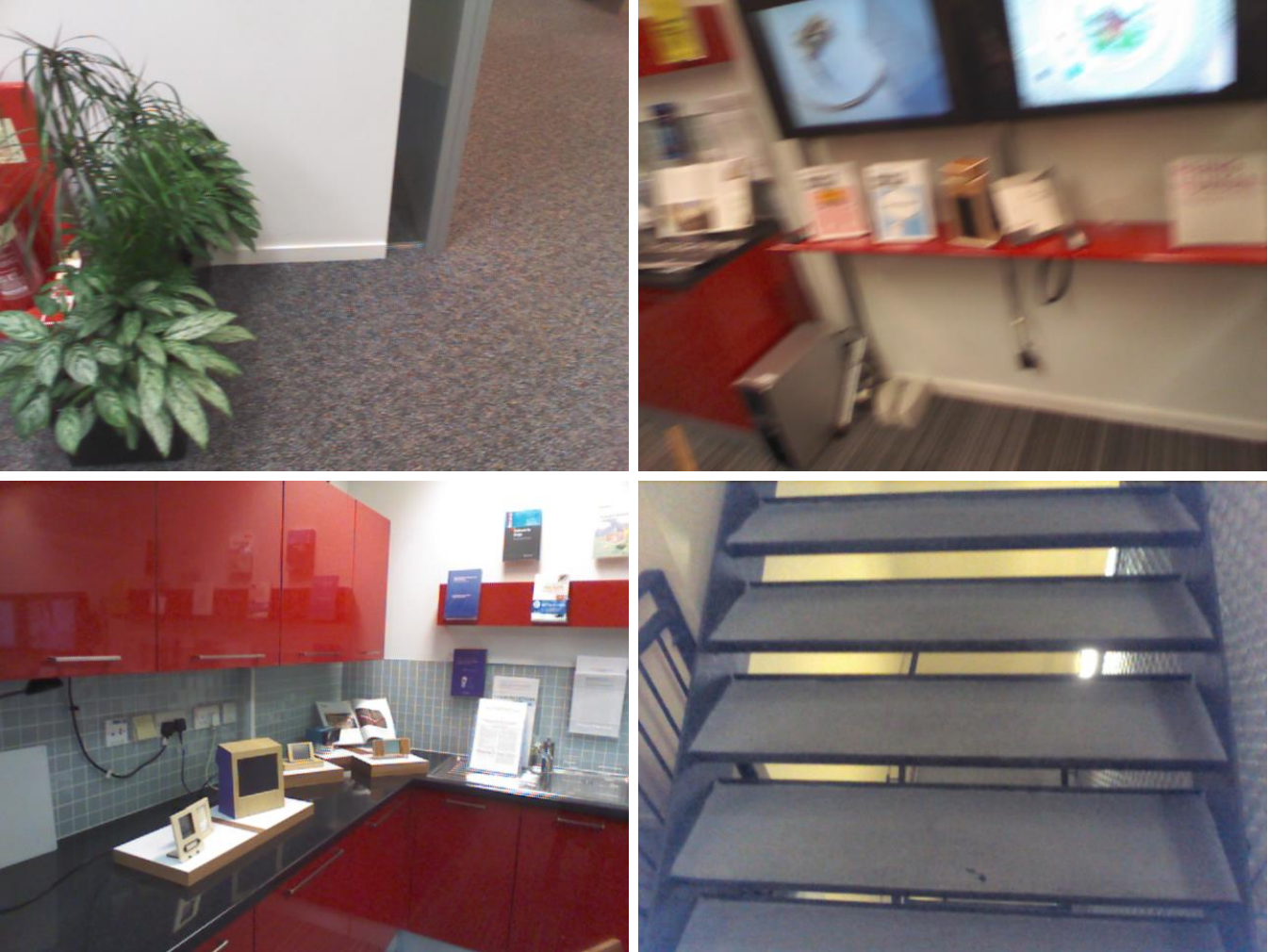}

   \caption{Difficult frames within the 7-Scenes dataset: Texture-less surfaces {\bf(upper left)}, motion blur {\bf(upper right)}, reflections {\bf(lower left)}, and repeating structures {\bf(lower right)}. DSAC estimates the correct pose in all 4 cases.}
\label{fig:7scenes:samples}
\end{figure}

\begin{figure}[h!t]

   \includegraphics[width=1.0\linewidth]{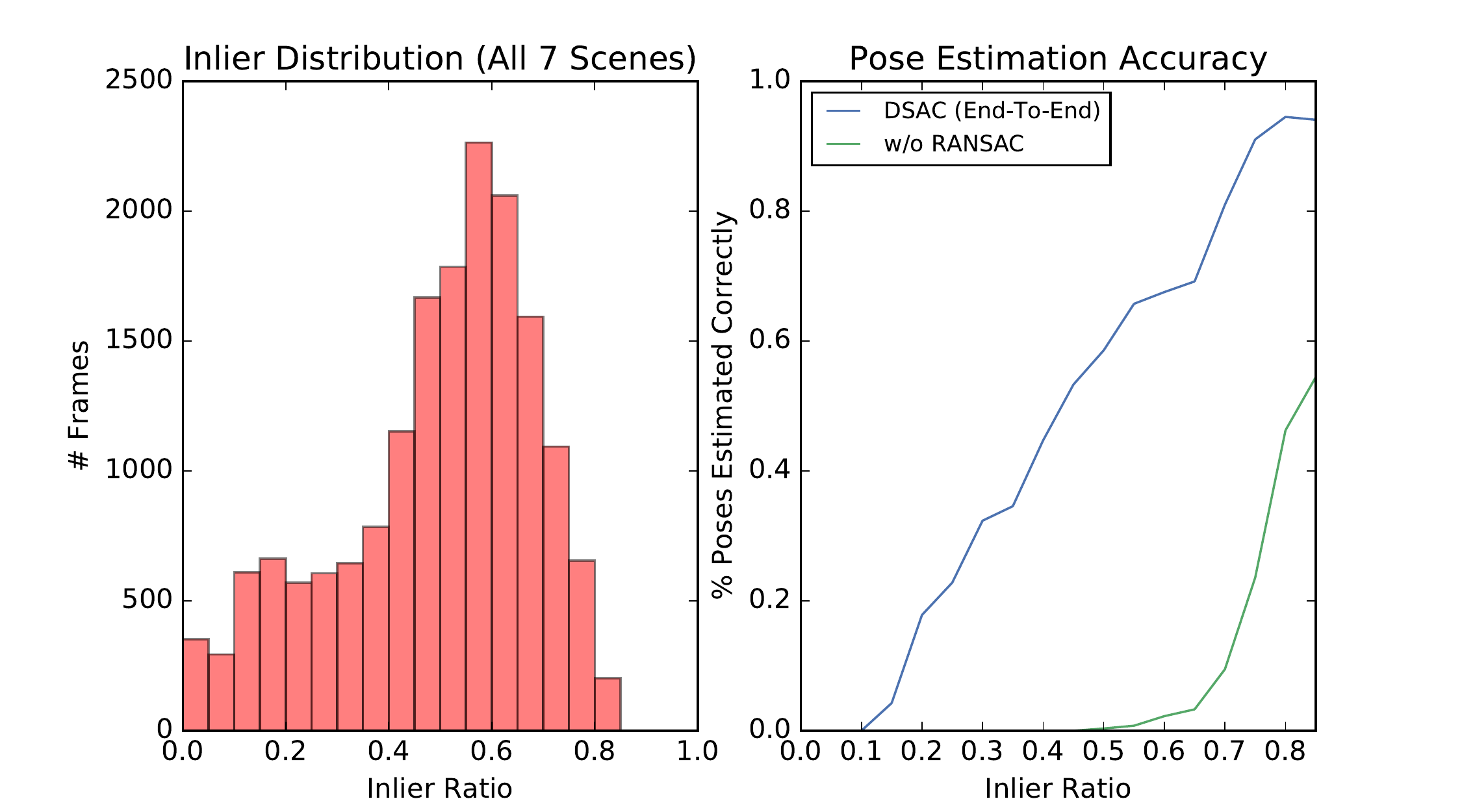}

   \caption{Distribution of inlier ratios of our scene coordinate predictions {\bf(left)}, and corresponding pose estimation accuracy of DSAC compared to a naive approach without RANSAC {\bf(right)}.}
\label{fig:inliers}
\end{figure}

Please see Fig.~\ref{fig:7scenes:samples} for examples of difficult situations in the 7-Scenes dataset. In our experiments, inlier ratios of scene coordinate predictions range from 5\% to 85\%. See Fig.~\ref{fig:inliers} (left) for the inlier ratio distribution over the complete 7-Scenes dataset. In accordance to \cite{shotton13scorf, brachmann2016}, we consider a scene coordinate prediction an inlier if it is within 10cm of the ground truth scene coordinate. In Fig.~\ref{fig:inliers} (right) we plot the performance of DSAC against the ratio of inliers. For comparison we plot the performance of a naive approach without RANSAC (pose fit to all scene coordinate predictions).

\section{Calculation of the Camera Pose Error}
\label{app:error}

An earlier version of this work differed in the exact numbers presented in Table \ref{exp:res:acc}. For example, our method scored approximately $56.8\%$ on the complete 7-Scenes set when trained componentwise. However, these numbers were produced with an error in the camera pose evaluation. In our formulation of the camera localization problem, we search for the pose $\mdl$ which aligns scene coordinate $\crd$ and their projections $\mathbf{p}$: $\mathbf{p} = C\mdl\crd$. However, in this formulation, $\mdl$ is \emph{not} the camera pose but the scene pose, \ie the \emph{inverse} camera pose. The calculation of the pose error (rotational and translation error) depends on whether it is calculated for $\mdl$ or $\mdl^{-1}$. For the camera pose, $\mdl^{-1}$, rotational errors contribute additionally to translational errors which, therefore, tend to be larger. Using the correct pose evaluation (\ie using $\mdl^{-1}$ to calculate rotational and translational errors) our results decreased to $45.5\%$. However, we found that the implementation of the PnP algorithm used in our experiments was a major limiting factor \wrt accuracy. Exchanging it for a standard, iterative PnP algorithm improved our results considerably, yielding the numbers we present in the current version of this work. Note that all conclusions drawn throughout the experimental section were valid for both, the original version and the updated version of this work.

{\small
\bibliographystyle{ieee}
\bibliography{diffransac}
}

\end{document}